\documentclass{article}
\usepackage[preprint]{neurips_2025}

\usepackage[utf8]{inputenc}
\usepackage{graphicx}
\usepackage{float}
\usepackage{amsmath,amssymb}
\usepackage{booktabs}
\usepackage{natbib}
\usepackage{hyperref}

\title{Preliminary Quantitative Study on Explainability and Trust in AI Systems}
\author{
  Allen Daniel Sunny \\
  University of Maryland, College Park \\
  \texttt{allens@umd.edu}
}

\begin{document}
\maketitle

% ----------------------------------------------------------------------
\begin{abstract}
Large-scale AI models such as GPT-4 have accelerated the deployment of artificial intelligence across critical domains including law, healthcare, and finance, raising urgent questions about trust and transparency. This study investigates the relationship between explainability and user trust in AI systems through a quantitative experimental design. Using an interactive, web-based loan approval simulation, we compare how different types of explanations---ranging from basic feature importance to interactive counterfactuals---influence perceived trust. Results suggest that interactivity enhances both user engagement and confidence, and that the clarity and relevance of explanations are key determinants of trust. These findings contribute empirical evidence to the growing field of human-centered explainable AI, highlighting measurable effects of explainability design on user perception.
\end{abstract}

% ----------------------------------------------------------------------
\section{Introduction}

\section{Introduction}

Artificial intelligence (AI) systems are now deeply embedded in critical areas of social and economic life; From credit scoring and hiring to healthcare triage and welfare eligibility. As these systems increasingly mediate access to opportunities and resources, they inherit profound questions of trust, accountability, and fairness. Although algorithmic decision-making promises efficiency and scale, its opacity has generated growing public concern over explainability and human oversight \citep{Raji2023, Whittlestone2022}. In domains where outcomes directly affect livelihoods, users often hesitate to rely on AI recommendations they cannot fully understand. Thus, fostering trust in AI systems is not only a technical challenge but also a social imperative.

Trust serves as the foundation for meaningful human-AI interaction. It shapes whether individuals accept, contest, or override algorithmic advice and determines how organizations adopt AI-driven tools. Prior research has emphasized two major routes toward trust: improving model accuracy and increasing system transparency \citep{Afroogh2024, Roszel2021}. Yet empirical studies consistently show that accuracy alone does not translate into trustworthiness, particularly when users are uncertain about how or why a model arrived at a given outcome. Explanations are therefore central to trust calibration, the process by which users align their confidence in a system with its actual reliability \citep{Hoffman2021}. The challenge lies not merely in generating explanations but in designing them to match user expectations, cognitive capacities, and goals.

Despite rapid progress in Explainable AI (XAI), many methods remain developer-centric, emphasizing mathematical fidelity over human interpretability. Feature-importance visualizations and textual rationales, while useful for diagnostics, often fail to provide the kind of contextual understanding that real users seek in decision-making environments. When an AI system denies a loan or recommends a medical procedure, individuals care less about feature weights and more about “why me?” or “what could I change?” These inherently contrastive and interactive questions underscore the need for explanations that mirror human reasoning patterns \citep{Ehsan2020, Miller2019, Le2023}. Counterfactual and interactive explanations offer a promising path forward, bridging algorithmic logic with human-centered inquiry.

To address this gap, our study conducts a quantitative investigation into how different types of explanations: ranging from basic feature-based to interactive counterfactual forms—affect user trust. Using a web-based loan approval simulation, participants engaged with AI systems of varying reliability under four explanation conditions: none, basic, contextual, and interactive. This design allows for a controlled comparison of how explanation style, model performance, and user expertise jointly shape perceptions of reliability and fairness. We hypothesize that interactive explanations will yield the highest trust and perceived understanding, while excessive detail may introduce cognitive fatigue and lower clarity.

By empirically examining these relationships, this work contributes measurable insights to the field of Human-Centered Explainable AI. Whereas prior research has relied heavily on qualitative studies or conceptual arguments, we provide quantitative evidence linking explanation design to user trust outcomes. The findings aim to inform both theory and practice—offering guidance for interface designers, policymakers, and AI developers seeking to implement trustworthy and transparent systems in real-world contexts.

% ----------------------------------------------------------------------
\section{Related Work}
The field of Explainable Artificial Intelligence (XAI) has evolved rapidly over the last decade, transitioning from early model-centric approaches toward user-focused interpretability frameworks. Initially, explainability research emphasized algorithmic transparency—developing models that could be interpreted by data scientists rather than end users. Classic examples include inherently interpretable models such as linear regressions, decision trees, and rule-based systems, which allowed developers to directly trace model logic. However, as predictive performance came to rely on high-capacity nonlinear models such as neural networks and ensemble methods, interpretability became increasingly opaque, motivating a wave of post-hoc techniques designed to explain complex models without altering their structure \citep{Ribeiro2016, Lundberg2017}.  

While these advances improved visibility into model mechanics, several scholars have argued that traditional XAI methods remain insufficient for promoting real-world trust and accountability \citep{Nguyen2024, Nauta2023}. The explanations they generate may satisfy a model developer’s curiosity but often fail to align with user needs, particularly in high-stakes applications where decisions must be justified to non-technical stakeholders. In response, the field has begun to pivot toward a human-centered perspective, emphasizing interpretability as a relational construct shaped by user context, cognitive capacity, and social meaning \citep{Ehsan2020, Miller2019}.  

\subsection{Local and Global Explanations}

One major distinction in the literature is between local and global explanations. \textbf{Global explanations} describe the overall behavior of a model—how it uses features on average across all predictions—and are useful for developers seeking to debug or audit algorithms \citep{Du2019}. \textbf{Local explanations}, in contrast, focus on individual instances, illuminating why a specific decision was made. Tools such as LIME \citep{Ribeiro2016} and SHAP \citep{Lundberg2017} provide localized feature attributions, helping users understand which variables most influenced a single output.  

Recent empirical studies indicate that users often find local explanations more intuitive because they map directly to personal outcomes rather than abstract model behavior \citep{Krause2016}. However, local explanations also have limitations: they can vary across similar cases and may overfit the explanation itself to local noise. This instability can paradoxically decrease user trust, particularly when explanations differ for near-identical inputs. Researchers such as \citet{Li2024} have proposed hybrid frameworks that combine global stability with local specificity to achieve consistency without sacrificing personalization. Our work builds on this line by using local explanations within a controlled decision-making setting where outcome consistency can be directly observed and measured through user trust ratings.

\subsection{Interactive and Counterfactual Methods}

Beyond static explanations, interaction has emerged as a critical determinant of explainability effectiveness. Interactive systems allow users to explore “what-if” scenarios, query the model’s reasoning, or visualize decision boundaries dynamically \citep{Yang2017, Fulton2020}. Such interactivity transforms explainability from a passive, one-directional disclosure into an active process of inquiry. Research on human-AI teaming suggests that interaction fosters a sense of agency and shared control, both of which are key predictors of trust \citep{Jacovi2021}.  

Counterfactual explanations occupy a related but distinct space in the XAI landscape. These explanations present alternative input conditions that would have led to a different decision—for example, “Had your income been \$5,000 higher, the loan would have been approved.” Because they align closely with human causal reasoning, counterfactuals offer intuitive and actionable insight \citep{Le2023}. They have also been linked to increased perceptions of fairness and empowerment, as users feel better informed about how to change outcomes. However, counterfactuals can introduce cognitive overload if presented without sufficient context or if the underlying model is inconsistent across similar examples.  

Several studies have begun combining interactivity and counterfactual reasoning to maximize transparency. For instance, \citet{Cai2019} and \citet{Nourani2019} show that allowing users to iteratively manipulate model inputs enhances understanding and long-term trust calibration. These hybrid techniques represent a shift from static post-hoc explanation to ongoing user-model dialogue. Our experiment builds directly on this paradigm, incorporating interactive counterfactuals within a gamified environment to observe how such features quantitatively affect user trust.

\subsection{Trust Calibration and Cognitive Factors}

Trust in AI is not a monolithic construct but a dynamic relationship between user perception and system reliability \citep{Hoffman2021, Thiebes2021}. Scholars distinguish between \textit{overtrust}, where users rely on AI excessively, and \textit{distrust}, where users underutilize accurate systems. Effective XAI should therefore aim to calibrate trust—ensuring that confidence levels align with actual model performance \citep{Schaefer2013}. Studies in human-automation interaction highlight several psychological dimensions that influence this calibration, including perceived competence, predictability, and value alignment \citep{Cahour2009, Roszel2021}.  

Cognitive factors such as explanation complexity and user expertise also play crucial roles. Novice users often benefit from simplified, narrative-style explanations, while experts prefer detailed technical breakdowns \citep{Ehsan2019}. When explanations exceed a user’s comprehension threshold, trust tends to decline regardless of accuracy. Conversely, overly simplistic explanations can appear patronizing or evasive, producing skepticism. Striking a balance between transparency and cognitive load is thus central to trustworthy AI design.  

In summary, existing literature identifies a clear need for empirical evidence linking explanation types to user trust outcomes. Most prior studies are qualitative or theoretical, leaving open questions about how different forms of explainability quantitatively shape user perceptions. This study directly addresses that gap by experimentally varying explanation modality and measuring corresponding shifts in perceived trust, reliability, and understanding across diverse user populations.

% ----------------------------------------------------------------------
\section{Methods}

\subsection{Experimental Design}
A web-based loan approval game was developed to evaluate explainability and trust. Each participant interacted with two AI models:
\begin{itemize}
    \item \textbf{Good AI:} A CatBoost classifier trained on the UCI Credit dataset \citep{Yeh2016Credit}, excluding demographic variables; accuracy $\approx$ 90\%.
    \item \textbf{Bad AI:} A flawed model with 40\% randomized targets, accuracy $\approx$ 65\%.
\end{itemize}

Participants reviewed 5 loan scenarios containing demographic and financial attributes (e.g., age, credit score, income) and received AI recommendations. Explanations were varied across four conditions:
\begin{enumerate}
    \item No explanation,
    \item Basic (feature importance),
    \item Detailed (contextual),
    \item Interactive (query-based).
\end{enumerate}

Each participant was assigned to one explainability condition and interacted with both AIs (order counterbalanced). The design followed a $3\times3\times2$ factorial structure across age, AI literacy, and system type.

\subsection{Participants}
Participants were recruited from university departments and nearby communities to ensure diverse representation. The final sample (N = 15) met power analysis requirements for medium effect sizes ($\alpha = .01, 1-\beta = .8$).  
Participants were categorized by:
\begin{itemize}
    \item \textbf{Age:} 18–25, 26–45, 46+,
    \item \textbf{AI familiarity:} novice, intermediate, expert.
\end{itemize}

\subsection{Trust and Explainability Measures}
\paragraph{Trust.}  
Trust was assessed via Likert-scale items adapted from \citet{Hoffman2021} and \citet{Cahour2009}, including confidence, predictability, reliability, accuracy, and efficiency.  
\paragraph{Explainability.}  
Explainability was evaluated using items derived from the COP-12 framework \citep{Nauta2023}, covering correctness, completeness, coherence, and contextual utility.  
Participants rated each item from 1 (strongly disagree) to 5 (strongly agree).

\subsection{Procedure}
After providing informed consent, participants completed a demographic and AI-literacy survey, followed by the loan-approval interaction. Each participant made decisions for 15 scenarios per AI system, after which they completed trust and explainability questionnaires. The web interface logged all decisions and explanation queries.

\begin{figure}[H]
\centering
\includegraphics[width=0.95\textwidth]{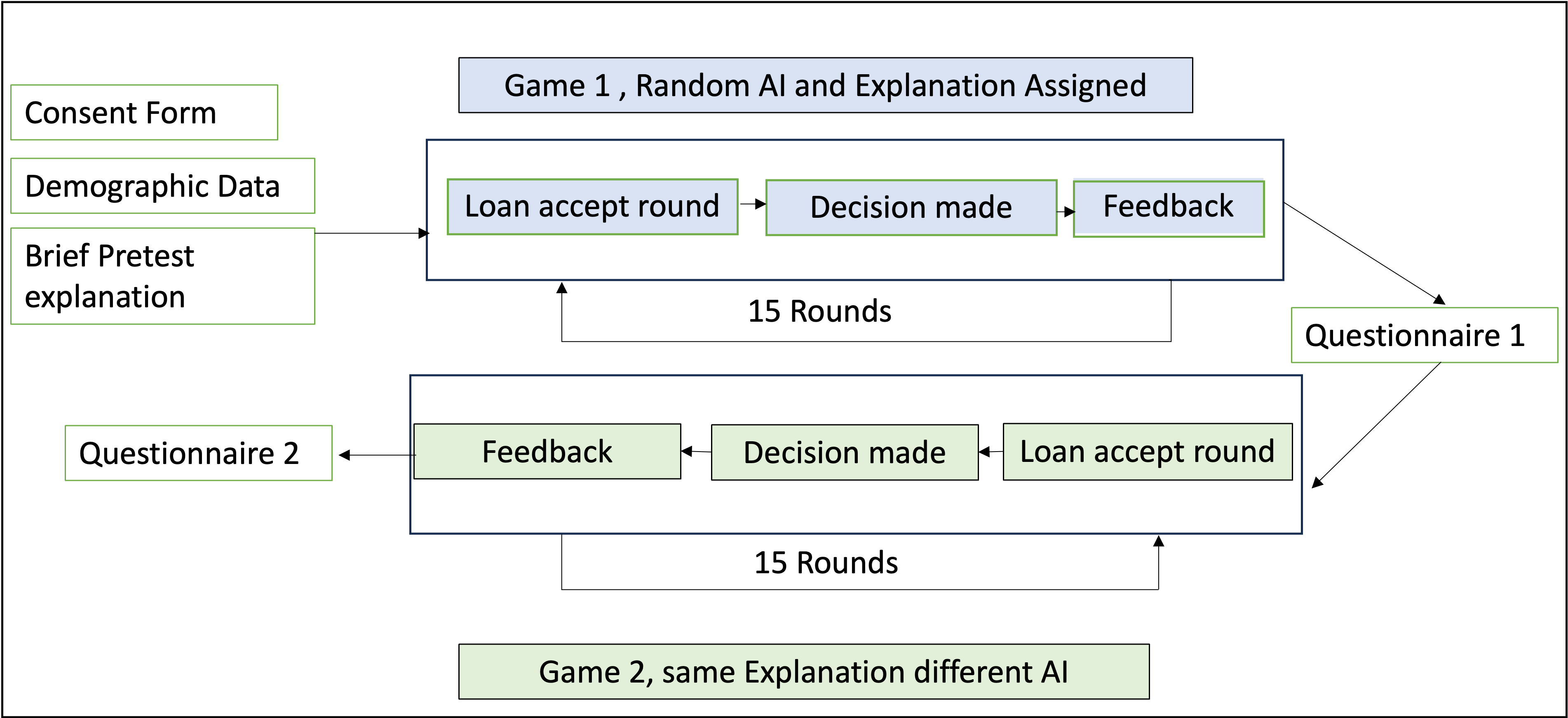}
\caption{Overview of study setup and participant flow.}
\label{fig:procedure}
\end{figure}

% ----------------------------------------------------------------------
\section{Results}

\subsection{Trust Outcomes}
Trust ratings differed significantly by explanation condition (ANOVA, $p<0.05$). Interactive explanations achieved the highest average trust (M = 4.22, SD = 0.61), followed by contextual (M = 3.87, SD = 0.58), basic (M = 3.51, SD = 0.62), and no explanation (M = 2.98, SD = 0.74).  

Participants using interactive systems also reported the lowest distrust and highest reliability confidence, particularly with the Good AI system. Novice users were most influenced by explanation presence, while experts exhibited calibrated trust across both AI systems.

\subsection{Explainability Ratings}
Explainability ratings mirrored trust trends: interactivity improved satisfaction and perceived detail but occasionally increased cognitive load. Participants preferred concise, actionable explanations over verbose technical ones.

\begin{table}[H]
\centering
\caption{Mean trust and explainability ratings by explanation type.}
\begin{tabular}{lcc}
\toprule
\textbf{Explanation Type} & \textbf{Mean Trust (1–5)} & \textbf{Mean Explainability (1–5)} \\
\midrule
None & 2.98 & 2.81 \\
Feature Importance & 3.51 & 3.42 \\
Contextual & 3.87 & 3.76 \\
Interactive & 4.22 & 4.15 \\
\bottomrule
\end{tabular}
\label{tab:trust}
\end{table}

% ----------------------------------------------------------------------
\section{Discussion}

Our results provide empirical evidence that both the form and the interactivity of AI explanations play a decisive role in shaping user trust. Participants who interacted with the \textit{interactive counterfactual} condition reported substantially higher trust and perceived understanding compared to those in static explanation conditions. This finding aligns with previous qualitative claims that explanation interactivity promotes engagement and agency \citep{Cai2019, Jacovi2021}. When users are allowed to query a model or explore “what-if” scenarios, they appear to treat the system less as an inscrutable oracle and more as a collaborative decision aid. Such agency supports a more stable trust calibration, confidence levels that reflect actual system competence rather than blind faith.

A key theme emerging from participant feedback was the tension between \textit{informativeness} and \textit{cognitive load}. While detailed explanations were valued for transparency, excessive textual or numerical information often led to fatigue and confusion. Participants repeatedly indicated a desire for succinct yet actionable rationales—explanations that communicate why an outcome occurred and how it might be changed, without unnecessary technical depth. This echoes prior research suggesting that the optimal explanation is not necessarily the most complete one, but the one that maximizes perceived clarity and relevance for the task at hand \citep{Ehsan2020}. From a design perspective, this highlights the importance of balancing explanation fidelity with cognitive ergonomics.

Another noteworthy observation involves the role of perceived fairness. Users frequently equated “understandable” decisions with “fair” ones. Even when exposed to the less accurate (``Bad AI'') system, participants expressed greater acceptance if they could interpret its reasoning or challenge it through interaction. This suggests that trust in AI systems extends beyond accuracy to encompass a moral dimension: people are more willing to tolerate model errors when they perceive the process as transparent and participatory. Such findings resonate with human–automation trust theory, which posits that procedural justice and perceived control strongly mediate trust outcomes \citep{Hoffman2021, Thiebes2021}.

Importantly, the quantitative results also support the notion that trust is context-dependent and dynamically constructed. The effect of explainability type varied significantly across user expertise levels. Expert participants tended to value detailed technical information and model reliability indicators, whereas novices responded more positively to narrative or example-based explanations. This divergence underscores the need for adaptive explainability—systems capable of tailoring the depth and presentation of explanations to user profiles. Adaptive or ``personalized'' XAI remains a nascent but promising direction for bridging expert and non-expert user needs \citep{Roszel2021, Nguyen2024}.

From a methodological standpoint, this study demonstrates the viability of using a controlled, gamified experiment to quantitatively evaluate trust in AI systems. The loan approval game framework offers ecological validity by simulating a realistic, high-stakes decision context while maintaining experimental control. The consistent patterns observed across demographic and competency groups reinforce the robustness of the findings. At the same time, these results should be interpreted as exploratory rather than definitive. The study’s moderate sample size and reliance on self-reported trust measures limit the generalizability of the conclusions. Future research could incorporate physiological or behavioral trust indicators , such as response latency, error correction behavior, or eye-tracking—to triangulate user confidence more objectively.

Another limitation involves the scope of explanation modalities tested. While the four explanation types capture a broad spectrum from non-interactive to highly interactive designs, real-world AI interfaces often blend multiple modalities (visual, textual, and numerical). Future work could explore multimodal explanations or longitudinal exposure to explanations over repeated interactions to assess trust persistence. Additionally, cross-cultural or policy-relevant applications—such as algorithmic decision-making in welfare or legal systems—could further illuminate how trust interacts with perceptions of legitimacy and authority.

In sum, this study contributes a quantitative foundation for understanding how explainability shapes user trust in AI. Interactive counterfactual explanations, by aligning with human causal reasoning and fostering user agency, represent a promising mechanism for building trustworthy AI systems. However, explanation design must remain sensitive to cognitive capacity, fairness perception, and contextual relevance. Trust cannot be imposed through transparency alone—it must be cultivated through iterative, human-centered design that treats explanation as dialogue rather than disclosure.

% ----------------------------------------------------------------------
\section{Limitations and Future Work}

This preliminary study relied on self-reported metrics within a simulated environment, limiting ecological validity. Future work will incorporate behavioral and physiological measures of trust, expand demographic diversity, and apply the framework to high-stakes domains such as healthcare and public benefit determination.  
Integrating legally grounded interpretability metrics could further align trust measurement with governance requirements.

% ----------------------------------------------------------------------
\section{Conclusion}

This paper presents quantitative evidence that interactive and contextual explanations significantly enhance user trust in AI systems. Beyond transparency, the results emphasize the role of user engagement and comprehension in shaping trustworthiness. These findings motivate future cross-disciplinary studies combining human-computer interaction, cognitive science, and responsible AI governance to develop scalable frameworks for trustworthy explainability.

Continued exploration in this direction could inform the design of adaptive explanation interfaces that adjust to user expertise, context, and decision stakes. Moreover, these insights call for the integration of participatory design and user feedback loops into AI system development, ensuring that explanations evolve alongside user expectations and sociotechnical contexts.

Future work should also investigate how explanation modalities, such as visual, textual, or interactive forms; Affect user understanding across diverse populations and decision domains. Expanding evaluation metrics beyond trust to include fairness, accountability, and satisfaction can provide a more holistic picture of responsible explainability.

Ultimately, bridging technical transparency with human-centered understanding will be essential for building AI systems that are not only accurate but meaningfully accountable to the people they serve. By grounding transparency in psychological realism and empirical usability, this work contributes to an ongoing shift toward explanation as a dialogue—one that empowers users, clarifies system boundaries, and reinforces the legitimacy of human oversight in algorithmic decision-making.% ----------------------------------------------------------------------

\appendix
\section{Appendix A: Trust Scale Used in the Study}

This appendix presents the complete trust measurement instrument used for evaluating participant
responses. The scale was administered using 5-point Likert items ranging from 1 (Strongly Disagree)
to 7 (Strongly Agree).

\begin{figure}[h]
    \centering
    \includegraphics[width=\linewidth]{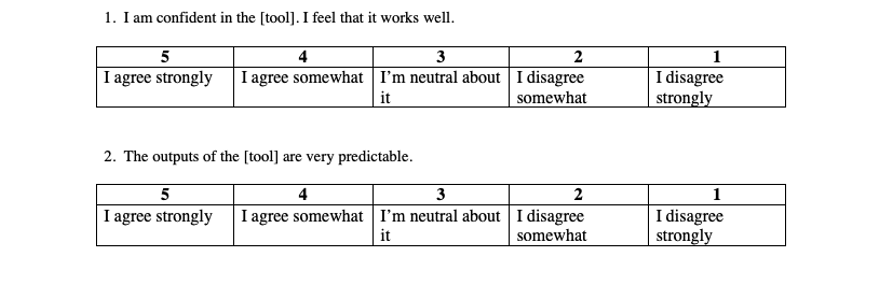}\\[0.5em]
    \includegraphics[width=\linewidth]{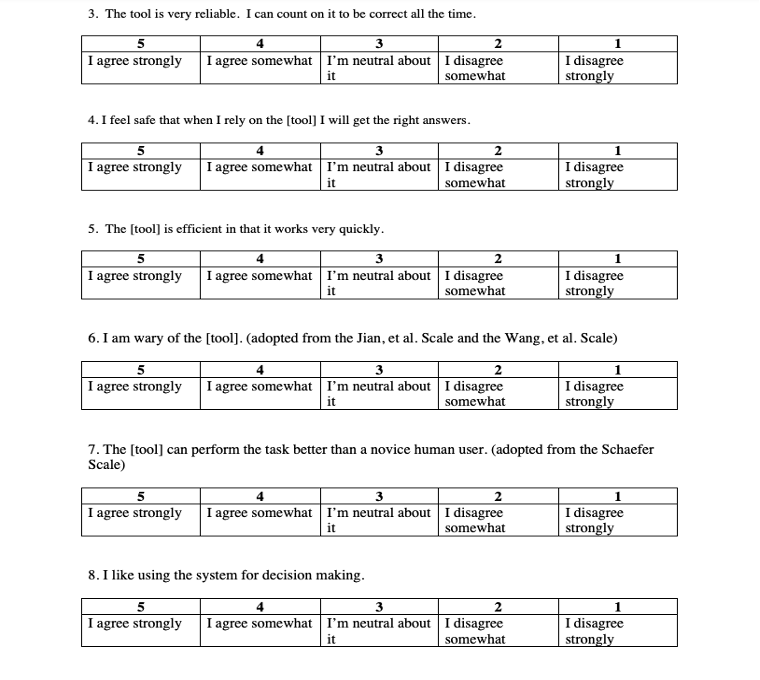}
    \caption{Full Trust Scale used in the study, presented across two sections for readability.}
    \label{fig:trust_scale_full}
\end{figure}

\vspace*{2em}

\section{Appendix B: Explainability Scale Used in the Study}

This appendix provides the explainability evaluation instrument adapted from the COP-12 framework
\citep{Nauta2023}. Participants rated each item on a 7-point Likert scale (1 = Strongly
Disagree, 7 = Strongly Agree).

\begin{figure}[h]
    \centering
    \includegraphics[width=\linewidth]{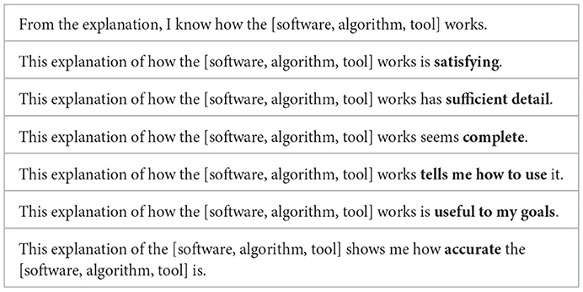}
    \caption{Explainability Scale used in the study. Items assess perceived correctness, completeness,
    coherence, and contextual utility of AI explanations.}
    \label{fig:explainability_scale}
\end{figure}

\vspace{1em}

\bibliographystyle{plainnat}
\bibliography{references}

\end{document}